\documentclass[10pt,journal,letterpaper,compsoc,twocolumn]{IEEEtran}

\usepackage{cite}
\usepackage{graphicx}
\usepackage{psfrag}
\usepackage{subfigure}
\usepackage{url}
\usepackage{stfloats}
\usepackage{amssymb,amsmath}
\usepackage{amsfonts}
\usepackage{array}
\usepackage{CJK}
\usepackage{setspace}
\usepackage[ruled]{algorithm2e}
\usepackage{cases}
\usepackage{listings}

\begin{document}

\title{Learning Topic Models by Belief Propagation}

\author{Jia~Zeng,~\IEEEmembership{Member,~IEEE,}
William~K.~Cheung,~\IEEEmembership{Member,~IEEE,}
and Jiming~Liu,~\IEEEmembership{Fellow,~IEEE}
\IEEEcompsocitemizethanks{\IEEEcompsocthanksitem
J.~Zeng is with the School of Computer Science and Technology,
Soochow University, Suzhou 215006, China.
He is also with the Shanghai Key Laboratory of Intelligent Information Processing, China.
To whom correspondence should be addressed.
E-mail: j.zeng@ieee.org.
\IEEEcompsocthanksitem
W.~K.~Cheung and J.~Liu are with the Department of Computer Science,
Hong Kong Baptist University, Kowloon Tong, Hong Kong.
}
}

\IEEEcompsoctitleabstractindextext{

\begin{abstract}
Latent Dirichlet allocation (LDA) is an important hierarchical Bayesian model for probabilistic topic modeling,
which attracts worldwide interests and touches on many important applications in text mining, computer vision and computational biology.
This paper represents LDA as a factor graph within the Markov random field (MRF) framework,
which enables the classic loopy belief propagation (BP) algorithm for approximate inference and parameter estimation.
Although two commonly-used approximate inference methods,
such as variational Bayes (VB) and collapsed Gibbs sampling (GS),
have gained great successes in learning LDA,
the proposed BP is competitive in both speed and accuracy as validated by encouraging experimental results on four large-scale document data sets.
Furthermore,
the BP algorithm has the potential to become a generic learning scheme for variants of LDA-based topic models.
To this end,
we show how to learn two typical variants of LDA-based topic models,
such as author-topic models (ATM) and relational topic models (RTM),
using BP based on the factor graph representation.
\end{abstract}


}

\maketitle

\IEEEdisplaynotcompsoctitleabstractindextext

%
\IEEEpeerreviewmaketitle

\section{Introduction} \label{s1}

Latent Dirichlet allocation (LDA)~\cite{Blei:03} is a three-layer hierarchical Bayesian model (HBM) that
can infer probabilistic word clusters called topics from the document-word (document-term) matrix.
LDA has no exact inference methods because of loops in its graphical representation.
Variational Bayes (VB)~\cite{Blei:03} and collapsed Gibbs sampling (GS)~\cite{Griffiths:04}
have been two commonly-used approximate inference methods for learning LDA and its extensions,
including author-topic models (ATM)~\cite{Rosen-Zvi:04} and relational topic models (RTM)~\cite{Chang:10}.
Other inference methods for probabilistic topic modeling include expectation-propagation (EP)~\cite{Minka:02}
and collapsed VB inference (CVB)~\cite{Teh:07}.
The connections and empirical comparisons among these approximate inference methods can be found in~\cite{Asuncion:09}.
Recently,
LDA and HBMs have found many important real-world applications in text mining and computer vision
(e.g.,
tracking historical topics from time-stamped documents~\cite{Iulian:10}
and activity perception in crowded and complicated scenes~\cite{WangXG:09}).

This paper represents LDA by the factor graph~\cite{Kschischang:01}
within the Markov random field (MRF) framework~\cite{Bishop:book}.
From the MRF perspective,
the topic modeling problem can be interpreted as a labeling problem,
in which the objective is to assign a set of semantic topic labels
to explain the observed nonzero elements in the document-word matrix.
MRF solves the labeling problem existing widely in image analysis and computer vision by two important concepts:
{\em neighborhood systems} and {\em clique potentials}~\cite{Zeng:08} or {\em factor functions}~\cite{Bishop:book}.
It assigns the best topic labels according to the {\em maximum a posteriori} (MAP) estimation
through maximizing the posterior probability,
which is in nature a prohibited combinatorial optimization problem in the discrete topic space.
However,
we often employ the smoothness prior~\cite{Li:book} over neighboring topic labels
to reduce the complexity by encouraging or penalizing only a limited number of possible labeling configurations.

The factor graph is a graphical representation method for both directed models (e.g., hidden Markov models (HMMs)~\cite[Chapter~13.2.3]{Bishop:book})
and undirected models (e.g., Markov random fields (MRFs)~\cite[Chapter~8.4.3]{Bishop:book}) because factor functions can represent both conditional and joint probabilities.
In this paper,
the proposed factor graph for LDA describes the same joint probability as that in the three-layer HBM,
and thus it is not a new topic model but interprets LDA from a novel MRF perspective.
The basic idea is inspired by the collapsed GS algorithm for LDA~\cite{Griffiths:04,Heinrich:08},
which integrates out multinomial parameters based on Dirichlet-Multinomial conjugacy
and views Dirichlet hyperparameters as the pseudo topic labels having the same layer with the latent topic labels.
In the collapsed hidden variable space,
the joint probability of LDA can be represented as the product of {\em factor functions} in the factor graph.
By contrast,
the undirected model ``harmonium"~\cite{Welling:04}
encodes a different joint probability from LDA and probabilistic latent semantic analysis (PLSA)~\cite{Hofmann:01},
so that it is a new and viable alternative to the directed models.

The factor graph representation facilitates the classic loopy belief propagation (BP) algorithm~\cite{Bishop:book,Kschischang:01,Frey:07}
for approximate inference and parameter estimation.
By designing proper {\em neighborhood system} and {\em factor functions},
we may encourage or penalize different local labeling configurations in the neighborhood system to realize the topic modeling goal.
The BP algorithm operates well on the factor graph,
and it has the potential to become a generic learning scheme for variants of LDA-based topic models.
For example,
we also extend the BP algorithm to learn ATM~\cite{Rosen-Zvi:04} and RTM~\cite{Chang:10} based on the factor graph representations.
Although the convergence of BP is not guaranteed on general graphs~\cite{Bishop:book},
it often converges and works well in real-world applications.

The factor graph of LDA also reveals some intrinsic relations between HBM and MRF.
HBM is a class of directed models within the Bayesian network framework~\cite{Heinrich:08},
which represents the causal or conditional dependencies of observed and hidden variables in the hierarchical manner
so that it is difficult to factorize the joint probability of hidden variables.
By contrast,
MRF can factorize the joint distribution of hidden variables into
the product of factor functions according to the Hammersley-Clifford theorem~\cite{Hammersley:71},
which facilitates the efficient BP algorithm for approximate inference and parameter estimation.
Although learning HBM often has difficulty in estimating parameters and inferring hidden variables due to the causal coupling effects,
the alternative factor graph representation as well as
the BP-based learning scheme may shed more light on faster and more accurate algorithms for HBM.

The remainder of this paper is organized as follows.
In Section~\ref{s2} we introduce the factor graph interpretation for LDA,
and derive the loopy BP algorithm for approximate inference and parameter estimation.
Moreover,
we discuss the intrinsic relations between BP and other state-of-the-art approximate inference algorithms.
Sections~\ref{s3} and~\ref{s4} present how to learn ATM and RTM using the BP algorithms.
Section~\ref{s5} validates the BP algorithm on four document data sets.
Finally,
Section~\ref{s6} draws conclusions and envisions future work.

\section{Belief Propagation for LDA} \label{s2}

The probabilistic topic modeling task is to assign a set of semantic topic labels,
$\mathbf{z} = \{z^k_{w,d}\}$,
to explain the observed nonzero elements in the document-word matrix
$\mathbf{x} = \{x_{w,d}\}$.
The notations $1 \le k \le K$ is the topic index,
$x_{w,d}$ is the number of word counts at the index $\{w,d\}$,
$1 \le w \le W$ and $1 \le d \le D$ are the word index in the vocabulary and the document index in the corpus.
Table~\ref{notation} summarizes some important notations.

\begin{figure}[t]
\centering
\includegraphics[width=1\linewidth]{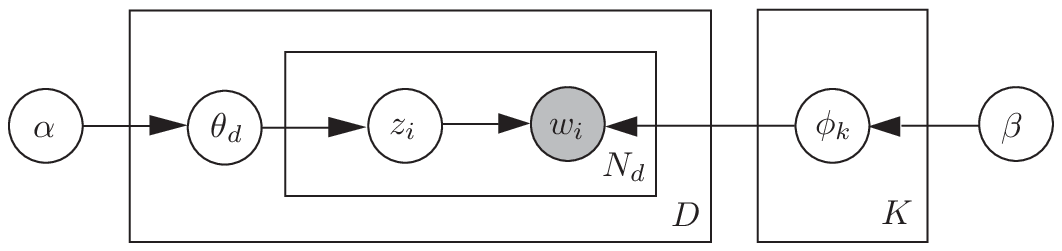}
\caption{Three-layer graphical representation of LDA~\cite{Blei:03}.
}
\label{lda}
\end{figure}

Fig.~\ref{lda} shows the original three-layer graphical representation of LDA~\cite{Blei:03}.
The document-specific topic proportion $\theta_d(k)$ generates a topic label $z^k_{w,d,i} \in \{0,1\}, \sum_{k=1}^K z^k_{w,d,i} = 1$,
which in turn generates each observed word token $i$ at the index $w$ in the document $d$
based on the topic-specific multinomial distribution $\phi_k(w)$ over the vocabulary words.
Both multinomial parameters $\theta_d(k)$ and $\phi_k(w)$ are generated by two Dirichlet distributions with hyperparameters $\alpha$ and $\beta$,
respectively.
For simplicity,
we consider only the smoothed LDA~\cite{Griffiths:04} with the fixed symmetric Dirichlet hyperparameters $\alpha$ and $\beta$.
The plates indicate replications.
For example,
the document $d$ repeats $D$ times in the corpus,
the word tokens $w_n$ repeats $N_d$ times in the document $d$,
the vocabulary size is $W$,
and there are $K$ topics.

\begin{table}[t]
\centering
\caption{Notations}
\begin{tabular}{|l|l|} \hline
$1 \le d \le D$                      &Document index              \\ \hline
$1 \le w \le W$                      &Word index in vocabulary    \\ \hline
$1 \le k \le K$                      &Topic index                 \\ \hline
$1 \le a \le A$                      &Author index                \\ \hline
$1 \le c \le C$                      &Link index                  \\ \hline
$\mathbf{x} = \{x_{w,d}\}$           &Document-word matrix \\ \hline
$\mathbf{z} = \{z^k_{w,d}\}$         &Topic labels for words      \\ \hline
$\mathbf{z}_{-w,d}$                  &Labels for $d$ excluding $w$   \\ \hline
$\mathbf{z}_{w,-d}$                  &Labels for $w$ excluding $d$   \\ \hline
$\mathbf{a}_d$                       &Coauthors of the document $d$ \\ \hline
$\boldsymbol{\mu}_{\cdot,d}(k)$      &$\sum_w x_{w,d}\mu_{w,d}(k)$       \\ \hline
$\boldsymbol{\mu}_{w,\cdot}(k)$      &$\sum_d x_{w,d}\mu_{w,d}(k)$       \\ \hline
$\theta_d$                           &Factor of the document $d$  \\ \hline
$\phi_w$                             &Factor of the word $w$      \\ \hline
$\eta_c$                             &Factor of the link $c$      \\ \hline
$f(\cdot)$                           &Factor functions            \\ \hline
$\alpha,\beta$                       &Dirichlet hyperparameters   \\ \hline
\end{tabular} \label{notation}
\end{table}

\subsection{Factor Graph Representation}

\begin{figure}[t]
\centering
\includegraphics[width=1\linewidth]{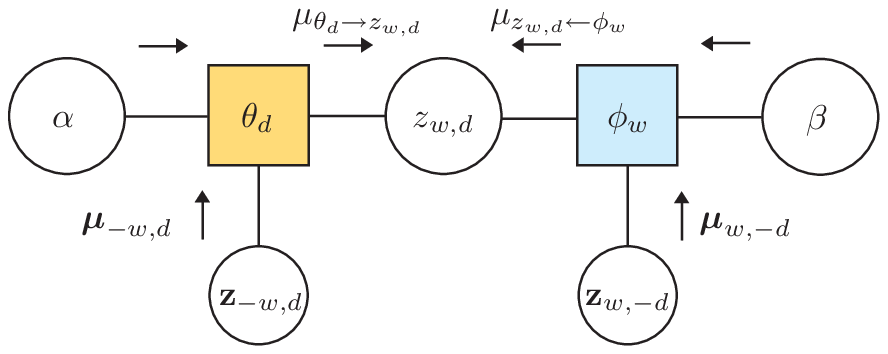}
\caption{Factor graph of LDA and message passing.
}
\label{ldapm}
\end{figure}

We begin by transforming the directed graph of Fig.~\ref{lda} into a two-layer factor graph,
of which a representative fragment is shown in Fig.~\ref{ldapm}.
The notation,
$z^k_{w,d} = \sum_{i=1}^{x_{w,d}} z^k_{w,d,i}/x_{w,d}$,
denotes the average topic labeling configuration
over all word tokens $1 \le i \le x_{w,d}$ at the index $\{w,d\}$.
We define the {\em neighborhood system} of the topic label $z_{w,d}$ as $\mathbf{z}_{-w,d}$ and $\mathbf{z}_{w,-d}$,
where $\mathbf{z}_{-w,d}$ denotes a set of topic labels associated with all word indices in the document $d$ except $w$,
and $\mathbf{z}_{w,-d}$ denotes a set of topic labels associated with the word indices $w$ in all documents except $d$.
The factors $\theta_d$ and $\phi_w$ are denoted by squares,
and their connected variables $z_{w,d}$ are denoted by circles.
The factor $\theta_d$ connects the neighboring topic labels $\{z_{w,d}, \mathbf{z}_{-w,d}\}$ at different word indices within the same document $d$,
while the factor $\phi_w$ connects the neighboring topic labels $\{z_{w,d}, \mathbf{z}_{w,-d}\}$ at the same word index $w$ but in different documents.
We absorb the observed word $w$ as the index of $\phi_w$,
which is similar to absorbing the observed document $d$ as the index of $\theta_d$.
Because the factors can be parameterized functions~\cite{Bishop:book},
both $\theta_d$ and $\phi_w$ can represent the same multinomial parameters with Fig.~\ref{lda}.

Fig.~\ref{ldapm} describes the same joint probability with Fig.~\ref{lda} if we properly design the factor functions.
The bipartite factor graph is inspired by the collapsed GS~\cite{Griffiths:04,Heinrich:08} algorithm,
which integrates out parameter variables $\{\theta, \phi\}$ in Fig.~\ref{lda}
and treats the hyperparameters $\{\alpha, \beta\}$ as pseudo topic counts having the same layer with hidden variables $\mathbf{z}$.
Thus,
the joint probability of the {\em collapsed} hidden variables can be factorized as the product of factor functions.
This collapsed view has been also discussed within the mean-field framework~\cite{Asuncion:10},
inspiring the zero-order approximation CVB (CVB0) algorithm~\cite{Asuncion:09} for LDA.
So,
we speculate that all three-layer LDA-based topic models can be collapsed into the two-layer factor graph,
which facilitates the BP algorithm for efficient inference and parameter estimation.
However,
how to use the two-layer factor graph to represent more general multi-layer HBM still remains to be studied.

Based on Dirichlet-Multinomial conjugacy,
integrating out multinomial parameters $\{\theta,\phi\}$ yields the joint probability~\cite{Heinrich:08} of LDA in Fig.~\ref{lda},
\begin{align} \label{ldaobj}
P(\mathbf{x},\mathbf{z}|\alpha,\beta) \propto
&\prod_{d=1}^D\prod_{k=1}^K \frac{\Gamma(\sum_{w=1}^W x_{w,d}z^k_{w,d} + \alpha)}{\Gamma[\sum_{k=1}^K(\sum_{w=1}^W x_{w,d}z^k_{w,d} + \alpha)]} \times \notag \\
\prod_{w=1}^W&\prod_{k=1}^K \frac{\Gamma(\sum_{d=1}^D x_{w,d}z^k_{w,d} + \beta)}{\Gamma[\sum_{w=1}^W(\sum_{d=1}^D x_{w,d}z^k_{w,d} + \beta)]},
\end{align}
where $x_{w,d}z^k_{w,d} = \sum_{i=1}^{x_{w,d}} z^k_{w,d,i}$ recovers the original topic configuration over the word tokens in Fig.~\ref{lda}.
Here,
we design the {\em factor functions} as
\begin{align}
f_{\theta_d}(\mathbf{x}_{\cdot,d},\mathbf{z}_{\cdot,d},\alpha) =
\prod_{k=1}^K \frac{\Gamma(\sum_{w=1}^W x_{w,d}z^k_{w,d} + \alpha)}{\Gamma[\sum_{k=1}^K(\sum_{w=1}^W x_{w,d}z^k_{w,d} + \alpha)]},  \\
f_{\phi_w}(\mathbf{x}_{w,\cdot},\mathbf{z}_{w,\cdot},\beta) =
\prod_{k=1}^K \frac{\Gamma(\sum_{d=1}^D x_{w,d}z^k_{w,d} + \beta)}{\Gamma[\sum_{w=1}^W(\sum_{d=1}^D x_{w,d}z^k_{w,d} + \beta)]},
\end{align}
where $\mathbf{z}_{\cdot,d} = \{z_{w,d},\mathbf{z}_{-w,d}\}$ and $\mathbf{z}_{w,\cdot} = \{z_{w,d},\mathbf{z}_{w,-d}\}$ denote subsets of the variables in Fig.~\ref{ldapm}.
Therefore,
the joint probability~\eqref{ldaobj} of LDA can be re-written as
the product of factor functions~\cite[Eq.~(8.59)]{Bishop:book} in Fig.~\ref{ldapm},
\begin{align}
P(\mathbf{x},\mathbf{z}|\alpha,\beta) \propto
\prod_{d=1}^{D} &f_{\theta_d}(\mathbf{x}_{\cdot,d},\mathbf{z}_{\cdot,d},\alpha)\prod_{w=1}^W f_{\phi_w}(\mathbf{x}_{w,\cdot},\mathbf{z}_{w,\cdot},\beta).
\end{align}
Therefore,
the two-layer factor graph in Fig.~\ref{ldapm} encodes exactly the same information with the three-layer graph for LDA in Fig.~\ref{lda}.
In this way,
we may interpret LDA within the MRF framework to treat probabilistic topic modeling as a labeling problem.

\subsection{Belief Propagation (BP)}

The BP~\cite{Bishop:book} algorithms provide exact solutions for
inference problems in tree-structured factor graphs but approximate solutions in factor graphs with loops.
Rather than directly computing the conditional joint probability $p(\mathbf{z}|\mathbf{x})$,
we compute the conditional marginal probability,
$p(z^k_{w,d}=1, x_{w,d}|\mathbf{z}^k_{-w,-d}, \mathbf{x}_{-w,-d})$,
referred to as {\em message} $\mu_{w,d}(k)$,
which can be normalized using a local computation, i.e.,
$\sum_{k=1}^K \mu_{w,d}(k) = 1, 0 \le \mu_{w,d}(k) \le 1$.
According to the Markov property in Fig.~\ref{ldapm},
we obtain
\begin{align} \label{marginal}
&p(z^k_{w,d}, x_{w,d}|\mathbf{z}^k_{-w,-d}, \mathbf{x}_{-w,-d}) \propto \notag \\
&p(z^k_{w,d}, x_{w,d}|\mathbf{z}^k_{-w,d}, \mathbf{x}_{-w,d})p(z^k_{w,d}, x_{w,d}|\mathbf{z}^k_{w,-d}, \mathbf{x}_{w,-d}),
\end{align}
where $-w$ and $-d$ denote all word indices except $w$ and all
document indices except $d$, and the notations $\mathbf{z}_{-w,d}$
and $\mathbf{z}_{w,-d}$ represent all possible neighboring topic
configurations. From the message passing view, $p(z^k_{w,d},
x_{w,d}|\mathbf{z}^k_{-w,d}, \mathbf{x}_{-w,d})$ is the neighboring
message $\mu_{\theta_d \rightarrow z_{w,d}}(k)$ sent from the factor
node $\theta_d$, and $p(z^k_{w,d}, x_{w,d}|\mathbf{z}^k_{w,-d},
\mathbf{x}_{w,-d})$ is the other neighboring message $\mu_{\phi_w
\rightarrow z_{w,d}}(k)$ sent from the factor node $\phi_w$. Notice
that~\eqref{marginal} uses the smoothness prior in MRF, which
encourages only $K$ smooth topic configurations within the
neighborhood system. Using the Bayes' rule and the joint
probability~\eqref{ldaobj}, we can expand Eq.~\eqref{marginal} as
\begin{align} \label{marginal2}
\mu_{w,d}(k)\propto&\frac{p(\mathbf{z}^k_{\cdot,d}, \mathbf{x}_{\cdot,d})}{p(\mathbf{z}^k_{-w,d}, \mathbf{x}_{-w,d})}
\times \frac{p(\mathbf{z}^k_{w,\cdot}, \mathbf{x}_{w,\cdot})}{p(\mathbf{z}^k_{w,-d}, \mathbf{x}_{w,-d})}, \notag \\
\propto&\frac{\sum_{-w} x_{-w,d}z^k_{-w,d} + \alpha}{\sum_{k=1}^K(\sum_{-w} x_{-w,d}z^k_{-w,d} + \alpha)} \times \notag \\
&\frac{\sum_{-d} x_{w,-d}z^k_{w,-d} + \beta}{\sum_{w=1}^W(\sum_{-d} x_{w,-d}z^k_{w,-d} + \beta)},
\end{align}
where the property,
$\Gamma(x+1)=x\Gamma(x)$,
is used to cancel the common terms in both nominator and denominator~\cite{Heinrich:08}.
We find that Eq.~\eqref{marginal2} updates the message on the variable $z^k_{w,d}$
if its neighboring topic configuration $\{\mathbf{z}^k_{-w,d}, \mathbf{z}^k_{w,-d}\}$ is known.
However,
due to uncertainty,
we know only the neighboring messages rather than the precise topic configuration.
So,
we replace topic configurations by corresponding messages in Eq.~\eqref{marginal2} and obtain the following message update equation,
\begin{align} \label{message}
\mu_{w,d}(k)\propto\frac{\boldsymbol{\mu}_{-w,d}(k) + \alpha}{\sum_k[\boldsymbol{\mu}_{-w,d}(k) + \alpha]} \times
&\frac{\boldsymbol{\mu}_{w,-d}(k) + \beta}{\sum_w[\boldsymbol{\mu}_{w,-d}(k) + \beta]},
\end{align}
where
\begin{gather}
\boldsymbol{\mu}_{-w,d}(k) = \sum_{-w} x_{-w,d}\mu_{-w,d}(k), \\
\boldsymbol{\mu}_{w,-d}(k) = \sum_{-d} x_{w,-d}\mu_{w,-d}(k).
\end{gather}

\begin{figure}[t]
\centering
\includegraphics[width=1\linewidth]{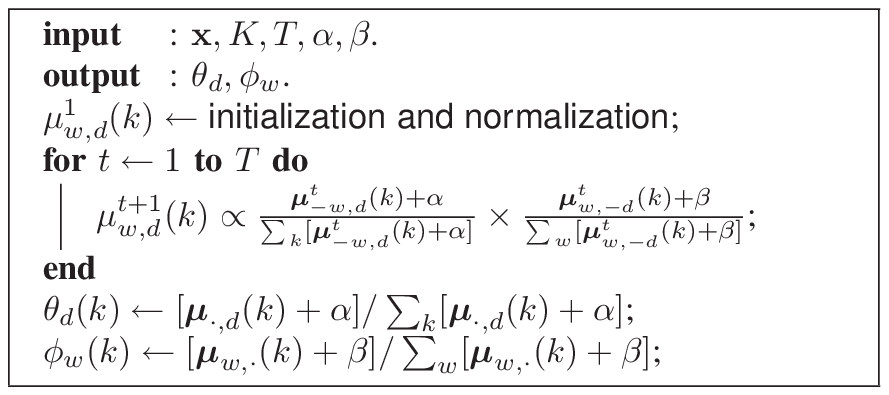}
\caption{The synchronous BP for LDA.}
\label{BP}
\end{figure}

Messages are passed from variables to factors,
and in turn from factors to variables until convergence or the maximum number of iterations is reached.
Notice that we need only pass messages for $x_{w,d} \ne 0$. Because $\mathbf{x}$ is a very sparse matrix,
the message update equation~\eqref{message} is computationally fast by sweeping all nonzero elements in the sparse matrix $\mathbf{x}$.

Based on messages,
we can estimate the multinomial parameters $\theta$ and $\phi$ by the expectation-maximization (EM) algorithm~\cite{Heinrich:08}.
The E-step infers the normalized message $\mu_{w,d}(k)$.
Using the Dirichlet-Multinomial conjugacy and Bayes' rule,
we express the marginal Dirichlet distributions on parameters as follows,
\begin{align}
\label{theta}
p(\theta_d) = \text{Dir}(\theta_d|\boldsymbol{\mu}_{\cdot,d}(k)+\alpha), \\
p(\phi_w) = \text{Dir}(\phi_w|\boldsymbol{\mu}_{w,\cdot}(k)+\beta).
\label{phi}
\end{align}
The M-step maximizes~\eqref{theta} and~\eqref{phi} with respect to $\theta_d$ and $\phi_w$,
resulting in the following point estimates of multinomial parameters,
\begin{gather}
\label{thetad}
\theta_d(k) = \frac{\boldsymbol{\mu}_{\cdot,d}(k) + \alpha}{\sum_k [\boldsymbol{\mu}_{\cdot,d}(k) + \alpha]}, \\
\phi_w(k) = \frac{\boldsymbol{\mu}_{w,\cdot}(k) + \beta}{\sum_w [\boldsymbol{\mu}_{w,\cdot}(k) + \beta]}.
\label{phiw}
\end{gather}
In this paper,
we consider only fixed hyperparameters $\{\alpha,\beta\}$.
Interested readers can figure out how to estimate hyperparameters based on inferred messages in~\cite{Heinrich:08}.

To implement the BP algorithm,
we must choose either the synchronous or the asynchronous update schedule to pass messages~\cite{Tappen:03}.
Fig.~\ref{BP} shows the synchronous message update schedule.
At each iteration $t$,
each variable uses the incoming messages in the previous iteration $t-1$ to calculate the current message.
Once every variable computes its message,
the message is passed to the neighboring variables and used to compute messages in the next iteration $t+1$.
An alternative is the asynchronous message update schedule.
It updates the message of each variable immediately.
The updated message is immediately used to compute other neighboring messages at each iteration $t$.
The asynchronous update schedule often passes messages faster across variables,
which causes the BP algorithm converge faster than the synchronous update schedule.
Another termination condition for convergence
is that the change of the multinomial parameters~\cite{Blei:03} is less than a predefined threshold $\lambda$,
for example,
$\lambda = 0.00001$~\cite{Hoffman:10}.

\subsection{An Alternative View of BP} \label{s2.3}

We may also adopt one of the BP instantiations, the sum-product algorithm~\cite{Bishop:book}, to infer $\mu_{w,d}(k)$.
For convenience,
we will not include the observation $x_{w,d}$ in the formulation.
Fig.~\ref{ldapm} shows the message passing from two factors $\theta_d$ and $\phi_w$ to the variable $z_{w,d}$,
where the arrows denote the message passing directions.
Based on the smoothness prior,
we encourage only $K$ smooth topic configurations without considering all other possible configurations.
The message $\mu_{w,d}(k)$ is proportional to the product of both incoming messages from factors,
\begin{align} \label{f2v}
\mu_{w,d}(k) \propto \mu_{\theta_d \rightarrow z_{w,d}}(k) \times \mu_{\phi_w \rightarrow z_{w,d}}(k).
\end{align}
Eq.~\eqref{f2v} has the same meaning with~\eqref{marginal}.
The messages from factors to variables are the sum of all incoming messages from the neighboring variables,
\begin{align} \label{v2f3}
\mu_{\theta_d \rightarrow z_{w,d}}(k) = f_{\theta_d} \prod_{-w} \mu_{-w,d}(k) \alpha, \\
\label{v2f4}
\mu_{\phi_w \rightarrow z_{w,d}}(k) = f_{\phi_w} \prod_{-d} \mu_{w,-d}(k) \beta,
\end{align}
where $\alpha$ and $\beta$ can be viewed as the pseudo-messages,
and $f_{\theta_d}$ and $f_{\phi_w}$ are the factor functions that encourage or penalize the incoming messages.

In practice,
however,
Eqs.~\eqref{v2f3} and~\eqref{v2f4} often cause the product of multiple incoming messages close to zero~\cite{Zeng:08}.
To avoid arithmetic underflow,
we use the sum operation rather than the product operation of incoming messages because when the product value increases the sum value also increases,
\begin{align}
\label{p2s1}
\prod_{-w} \mu_{-w,d}(k) \alpha \propto \sum_{-w} \mu_{-w,d}(k) + \alpha, \\
\label{p2s2}
\prod_{-d} \mu_{w,-d}(k) \beta \propto \sum_{-d} \mu_{w,-d}(k) + \beta.
\end{align}
Such approximations as~\eqref{p2s1} and~\eqref{p2s2} transform the sum-product to the sum-sum algorithm,
which resembles the relaxation labeling algorithm for learning MRF with good performance~\cite{Zeng:08}.

The normalized message $\mu_{w,d}(k)$ is multiplied by the number of word counts $x_{w,d}$ during the propagation,
i.e.,
$x_{w,d}\mu_{w,d}(k)$.
In this sense,
$x_{w,d}$ can be viewed as the weight of $\mu_{w,d}(k)$ during the propagation,
where the bigger $x_{w,d}$ corresponds to the larger influence of its message to those of its neighbors.
Thus,
the topics may be distorted by those documents with greater word counts.
To avoid this problem,
we may choose another weight like term frequency (TF) or
term frequency-inverse document frequency (TF-IDF) for {\em weighted} belief propagation.
In this sense,
BP can not only handle {\em discrete} data,
but also process {\em continuous} data like TF-IDF.
The MRF model in Fig.~\ref{ldapm} can be extended to describe both discrete and continuous data in general,
while LDA in Fig.~\ref{lda} focuses only on generating discrete data.

In the MRF model,
we can design the factor functions arbitrarily to encourage or penalize local topic labeling configurations based on our prior knowledge.
From Fig.~\ref{lda},
LDA solves the topic modeling problem according to three intrinsic assumptions:
\begin{enumerate}
\item
Co-occurrence: Different word indices within the same document tend to be associated with the same topic labels.
\item
Smoothness: The same word indices in different documents are likely to be associated with the same topic labels.
\item
Clustering: All word indices do not tend to associate with the same topic labels.
\end{enumerate}
The first assumption is determined by the document-specific topic proportion $\theta_d(k)$,
where it is more likely to assign a topic label $z^k_{w,d} = 1$ to the word index $w$ if the topic $k$ is more frequently assigned to other word indices in the document $d$.
Similarly,
the second assumption is based on the topic-specific multinomial distribution $\phi_k(w)$.
which implies a higher likelihood to associate the word index $w$ with the topic label $z^k_{w,d} = 1$ if $k$
is more frequently assigned to the same word index $w$ in other documents except $d$.
The third assumption avoids grouping all word indices into one topic through normalizing $\phi_k(w)$ in terms of all word indices.
If most word indices are associated with the topic $k$,
the multinomial parameter $\phi_k$ will become too small to allocate the topic $k$ to these word indices.

According to the above assumptions,
we design $f_{\theta_d}$ and  $f_{\phi_w}$ over messages as
\begin{align} \label{ftheta}
f_{\theta_d}(\boldsymbol{\mu}_{\cdot,d}, \alpha) = \frac{1}{\sum_k [\boldsymbol{\mu}_{-w,d}(k) + \alpha]}, \\
\label{fphi}
f_{\phi_w}(\boldsymbol{\mu}_{w,\cdot}, \beta) = \frac{1}{\sum_w [\boldsymbol{\mu}_{w,-d}(k) + \beta]}.
\end{align}
Eq.~\eqref{ftheta} normalizes the incoming messages by the total
number of messages for all topics associated with the document $d$
to make outgoing messages comparable across documents.
Eq.~\eqref{fphi} normalizes the incoming messages by the total
number of messages for all words in the vocabulary to make outgoing
messages comparable across vocabulary words.
Notice that~\eqref{v2f3} and~\eqref{v2f4} realize the first two assumptions,
and~\eqref{fphi} encodes the third assumption of topic modeling.
The similar normalization technique to avoid partitioning all data points into one cluster
has been used in the classic normalized cuts algorithm for image segmentation~\cite{Shi:00}.
Combining~\eqref{f2v} to \eqref{fphi} will yield the same message update equation~\eqref{message}.
To estimate parameters $\theta_d$ and $\phi_w$,
we use the joint marginal distributions~\eqref{v2f3} and~\eqref{v2f4} of the set of variables
belonging to factors $\theta_d$ and $\phi_w$ including the variable $z_{w,d}$,
which produce the same point estimation equations~\eqref{thetad} and~\eqref{phiw}.

\subsection{Simplified BP (siBP)} \label{sbp}

\begin{figure}[t]
\centering
\includegraphics[width=1\linewidth]{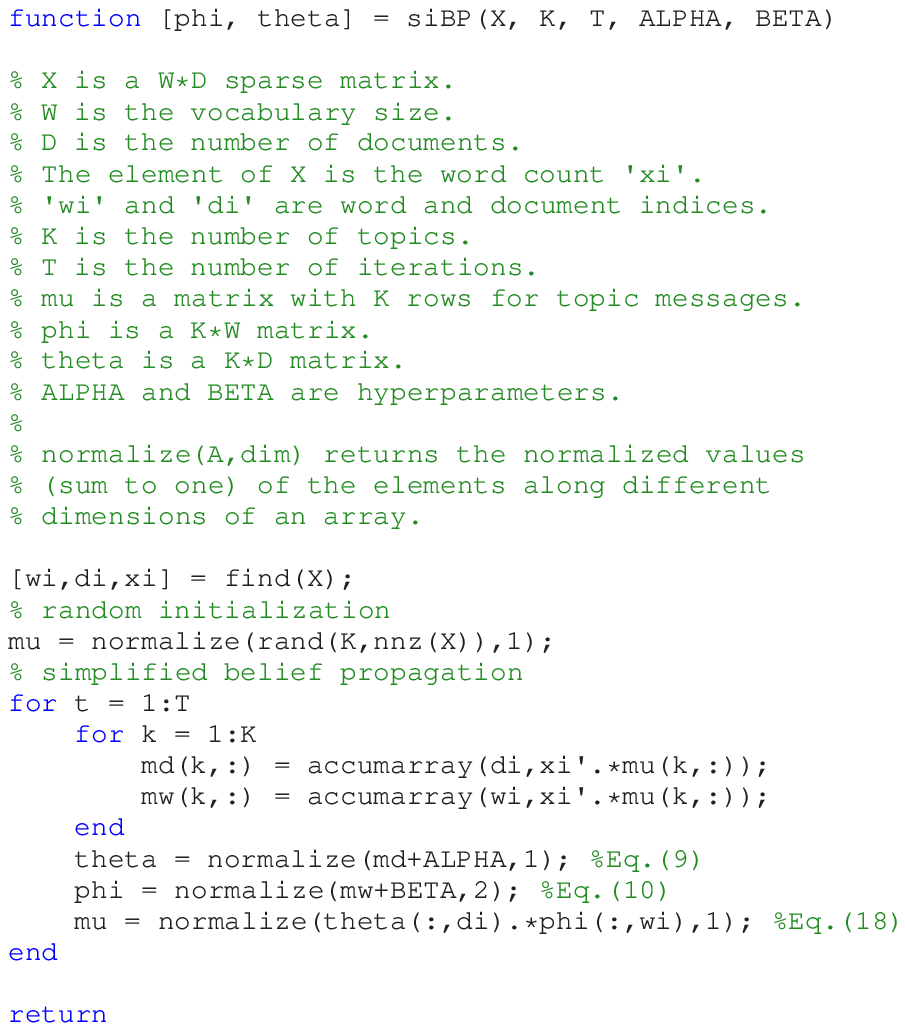}
\caption{The MATLAB code for siBP.
}
\label{siBP}
\end{figure}

We may simplify the message update equation~\eqref{message}.
Substituting~\eqref{thetad} and~\eqref{phiw} into~\eqref{message}
yields the approximate message update equation,
\begin{align} \label{approx}
\mu_{w,d}(k) \propto \theta_d(k) \times \phi_w(k),
\end{align}
which includes the current message $\mu_{w,d}(k)$ in both numerator and denominator in~\eqref{message}.
In many real-world topic modeling tasks,
a document often contains many different word indices,
and the same word index appears in many different documents.
So,
at each iteration,
Eq.~\eqref{approx} deviates slightly from~\eqref{message} after adding the current message to both numerator and denominator.
Such slight difference may be enlarged after many iterations in Fig.~\ref{BP} due to accumulation effects,
leading to different estimated parameters.
Intuitively,
Eq.~\eqref{approx} implies that if the topic $k$ has a higher proportion in the document $d$,
and it has the a higher likelihood to generate the word index $w$,
it is more likely to allocate the topic $k$ to the observed word $x_{w,d}$.
This allocation scheme in principle follows the three intrinsic topic modeling assumptions in the subsection~\ref{s2.3}.
Fig.~\ref{siBP} shows the MATLAB code for the simplified BP (siBP).

\subsection{Relationship to Previous Algorithms} \label{discussion}

Here we discuss some intrinsic relations between BP with three state-of-the-art LDA learning algorithms
such as VB~\cite{Blei:03}, GS~\cite{Griffiths:04},
and zero-order approximation CVB (CVB0)~\cite{Asuncion:09,Asuncion:10} within the unified message passing framework.
The message update scheme is an instantiation of the E-step of EM algorithm~\cite{Dempster:77},
which has been widely used to infer the marginal probabilities of hidden variables
in various graphical models according to the maximum-likelihood estimation~\cite{Bishop:book}
(e.g., the E-step inference for GMMs~\cite{Zeng:08c},
the forward-backward algorithm for HMMs~\cite{Zeng:06a},
and the probabilistic relaxation labeling algorithm for MRF~\cite{Zeng:08a}).
After the E-step,
we estimate the optimal parameters using the updated messages and observations at the M-step of EM algorithm.

VB is a variational message passing method~\cite{Winn:05} that
uses a set of factorized variational distributions $q(\mathbf{z})$
to approximate the joint distribution~\eqref{ldaobj} by minimizing the Kullback-Leibler (KL) divergence between them.
Employing the Jensen's inequality makes
the approximate variational distribution an adjustable lower bound on the joint distribution,
so that maximizing the joint probability is equivalent to maximizing the lower bound by tuning a set of variational parameters.
The lower bound $q(\mathbf{z})$ is also an MRF in nature that approximates the joint distribution~\eqref{ldaobj}.
Because there is always a gap between the lower bound and the true joint distribution,
VB introduces bias when learning LDA.
The variational message update equation is
\begin{align} \label{vb}
\mu_{w,d}(k) \propto \frac{\exp[\Psi(\boldsymbol{\mu}_{\cdot,d}(k) + \alpha)]}{\exp[\Psi(\sum_k [\boldsymbol{\mu}_{\cdot,d}(k) + \alpha])]} \times \notag \\
\frac{\boldsymbol{\mu}_{w,\cdot}(k) + \beta}{\sum_w [\boldsymbol{\mu}_{w,\cdot}(k) + \beta]},
\end{align}
which resembles the synchronous BP~\eqref{message} but with two major differences.
First,
VB uses complicated digamma functions $\Psi(\cdot)$,
which not only introduces bias~\cite{Asuncion:09} but also slows down the message updating.
Second,
VB uses a different variational EM schedule.
At the E-step,
it simultaneously updates both variational messages and parameter of $\theta_d$ until convergence,
holding the variational parameter of $\phi$ fixed.
At the M-step,
VB updates only the variational parameter of $\phi$.

The message update equation of GS is
\begin{align} \label{gs}
\mu_{w,d,i}(k) \propto \frac{n^{-i}_{\cdot,d}(k) + \alpha}{\sum_k [n^{-i}_{\cdot,d}(k) + \alpha]} \times
\frac{n^{-i}_{w,\cdot}(k) + \beta}{\sum_w [n^{-i}_{w,\cdot}(k) + \beta]},
\end{align}
where $n^{-i}_{\cdot,d}(k)$ is the total number of topic labels $k$ in the document $d$ except the topic label on the current word token $i$,
and $n^{-i}_{w,\cdot}(k)$ is the total number of topic labels $k$ of the word $w$ except the topic label on the current word token $i$.
Eq.~\eqref{gs} resembles the asynchronous BP implementation~\eqref{message} but with two subtle differences.
First,
GS randomly samples the current topic label $z^{k}_{w,d,i} = 1$ from the message $\mu_{w,d,i}(k)$,
which truncates all $K$-tuple message values to zeros except the sampled topic label $k$.
Such information loss introduces bias when learning LDA.
Second,
GS must sample a topic label for each word token,
which repeats $x_{w,d}$ times for the word index $\{w,d\}$.
The sweep of the entire word tokens rather than word index restricts GS's scalability to
large-scale document repositories containing billions of word tokens.

CVB0 is exactly equivalent to our asynchronous BP implementation but based on word tokens.
Previous empirical comparisons~\cite{Asuncion:09} advocated the CVB0 algorithm for LDA
within the approximate mean-field framework~\cite{Asuncion:10} closely connected with the proposed BP.
Here we clearly explain that the superior performance of CVB0
has been largely attributed to its asynchronous BP implementation from the MRF perspective.
Our experiments also support that the message passing over word indices instead of tokens
will produce comparable or even better topic modeling performance but with significantly smaller computational costs.

Eq.~\eqref{approx} also reveals that siBP is a probabilistic matrix factorization algorithm
that factorizes the document-word matrix,
$\mathbf{x} = [x_{w,d}]_{W \times D}$,
into a matrix of document-specific topic proportions,
$\boldsymbol{\theta} = [\theta_d(k)]_{K \times D}$,
and a matrix of vocabulary word-specific topic proportions,
$\boldsymbol{\phi} = [\phi_w(k)]_{K \times W}$,
i.e.,
$\mathbf{x} \sim \boldsymbol{\phi}^{\mathrm{T}}\boldsymbol{\theta}$.
We see that the larger number of word counts $x_{w,d}$
corresponds to the higher likelihood $\sum_k\theta_d(k)\phi_w(k)$.
From this point of view,
the multinomial principle component analysis (PCA)~\cite{Buntine:02} describes some intrinsic relations among LDA,
PLSA~\cite{Hofmann:01},
and non-negative matrix factorization (NMF)~\cite{Lee:99}.
Eq.~\eqref{approx} is the same as the E-step update for PLSA
except that the parameters $\theta$ and $\phi$ are smoothed by the hyperparameters $\alpha$ and $\beta$ to prevent overfitting.

VB, BP and siBP have the computational complexity $\mathcal{O}(KDW_dT)$,
but GS and CVB0 require $\mathcal{O}(KDN_dT)$,
where $W_d$ is the average vocabulary size,
$N_d$ is the average number of word tokens per document,
and $T$ is the number of learning iterations.

\section{Belief Propagation for ATM} \label{s3}

\begin{figure*}[t]
\centering
\includegraphics[width=0.7\linewidth]{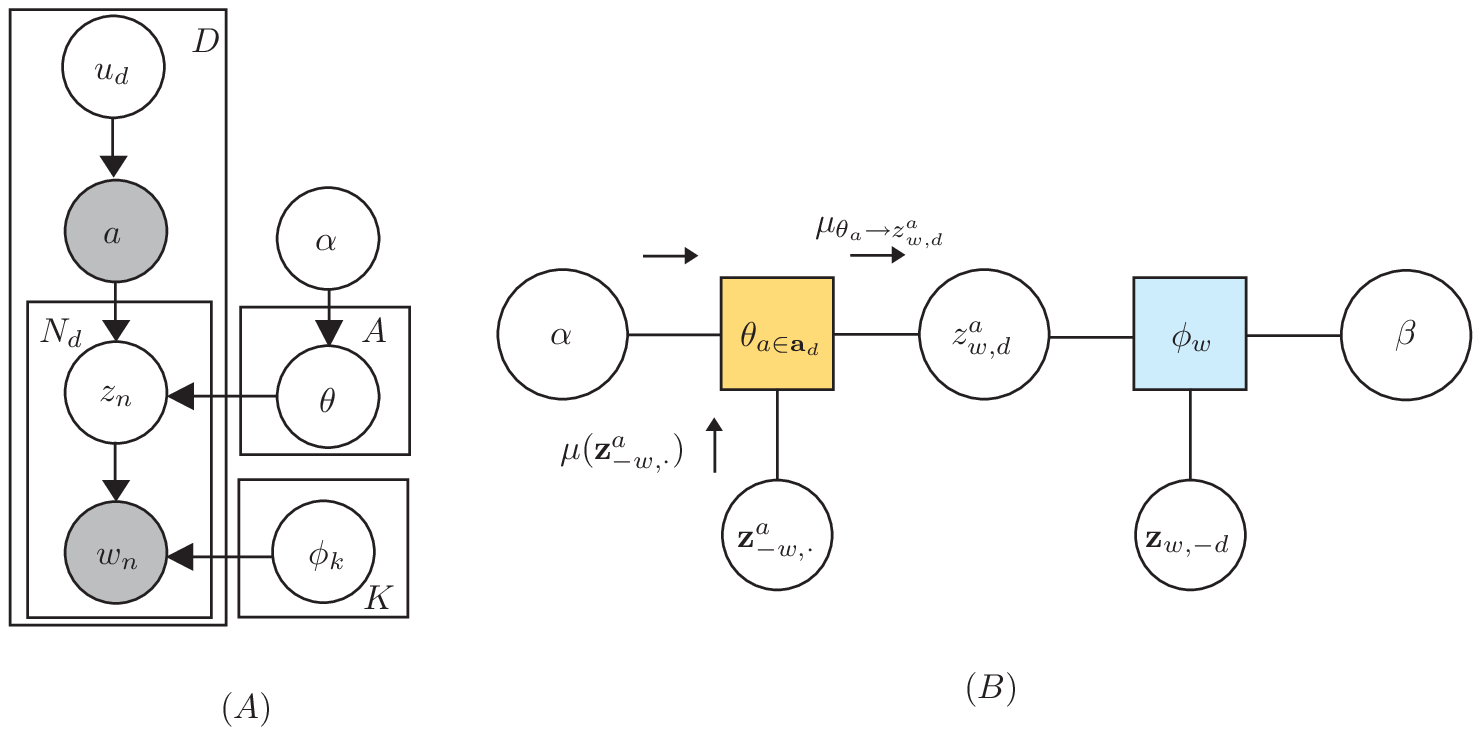}
\caption{(A) The three-layer graphical representation~\cite{Rosen-Zvi:04} and (B) two-layer factor graph of ATM.}
\label{atm}
\end{figure*}

Author-topic models (ATM)~\cite{Rosen-Zvi:04} depict each author of the document as a mixture of probabilistic topics,
and have found important applications in matching papers with reviewers~\cite{Zeng:10}.
Fig.~\ref{atm}A shows the generative graphical representation for ATM,
which first uses a document-specific uniform distribution $u_d$ to generate an author index $a, 1 \le a \le A$,
and then uses the author-specific topic proportions $\theta_a$ to generate a topic label $z^k_{w,d} = 1$ for the word index $w$ in the document $d$.
The plate on $\theta$ indicates that there are $A$ unique authors in the corpus.
The document often has multiple coauthors.
ATM randomly assigns one of the observed author indices to each word in the document based on the document-specific uniform distribution $u_d$.
However,
it is more reasonable that each word $x_{w,d}$ is associated with an author index $a \in \mathbf{a}_d$ from the multinomial rather than uniform distribution,
where $\mathbf{a}_d$ is a set of author indices of the document $d$.
As a result,
each topic label takes two variables $z^{a,k}_{w,d} = \{0, 1\}, \sum_{a,k}z^{a,k}_{w,d} = 1, a \in \mathbf{a}_d, 1 \le k \le K$,
where $a$ is the author index and $k$ is the topic index attached to the word.

We transform Fig.~\ref{atm}A to the factor graph representation of ATM in Fig.~\ref{atm}B.
As with Fig.~\ref{ldapm},
we absorb the observed author index $a \in \mathbf{a}_d$ of the document $d$ as the index of the factor $\theta_{a \in \mathbf{a}_d}$.
The notation $\mathbf{z}^a_{-w,\cdot}$ denotes all labels connected with the authors $a \in \mathbf{a}_d$ except those for the word index $w$.
The only difference between ATM and LDA is that the author $a \in \mathbf{a}_d$ instead of the document $d$ connects the labels $z^a_{w,d}$ and $\mathbf{z}^a_{-w,\cdot}$.
As a result,
ATM encourages topic smoothness among labels $z^a_{w,d}$ attached to the same author $a$ instead of the same document $d$.

\subsection{Inference and Parameter Estimation}

\begin{figure}[t]
\centering
\includegraphics[width=1\linewidth]{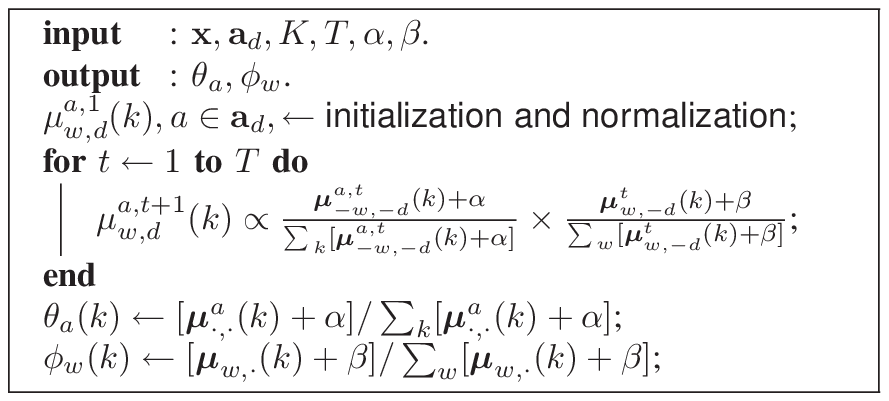}
\caption{The synchronous BP for ATM.}
\label{BPATM}
\end{figure}

Unlike passing the $K$-tuple message $\mu_{w,d}(k)$ in Fig.~\ref{BP},
the BP algorithm for learning ATM passes the $|\mathbf{a}_d|\times K$-tuple message vectors
$\mu^a_{w,d}(k), a \in \mathbf{a}_d$ through the factor $\theta_{a \in \mathbf{a}_d}$ in Fig.~\ref{atm}B,
where $|\mathbf{a}_d|$ is the number of authors in the document $d$.
Nevertheless,
we can still obtain the $K$-tuple word topic message $\mu_{w,d}(k)$
by marginalizing the message $\mu^a_{w,d}(k)$ in terms of the author variable $a \in \mathbf{a}_d$ as follows,
\begin{align} \label{wordmessage}
\mu_{w,d}(k) = \sum_{a \in \mathbf{a}_d} \mu^a_{w,d}(k).
\end{align}

Since Figs.~\ref{ldapm} and~\ref{atm}B have the same right half part,
the message passing equation from the factor $\phi_w$ to the variable $z_{w,d}$ and the parameter estimation equation for $\phi_w$
in Fig.~\ref{atm}B remain the same as~\eqref{message} and~\eqref{phiw}
based on the marginalized word topic message in~\eqref{wordmessage}.
Thus,
we only need to derive the message passing equation from the factor $\theta_{a \in \mathbf{a}_d}$ to the variable $z^a_{w,d}$ in Fig.~\ref{atm}B.
Because of the topic smoothness prior,
we design the factor function as follows,
\begin{align} \label{atmftheta}
f_{\theta_a} = \frac{1}{\sum_k [\boldsymbol{\mu}^a_{-w,-d}(k) + \alpha]},
\end{align}
where $\boldsymbol{\mu}^a_{-w,-d}(k) = \sum_{-w,-d} x^a_{w,d}\mu^a_{w,d}(k)$
denotes the sum of all incoming messages attached to the author index $a$ and the topic index $k$ excluding $x^a_{w,d}\mu^a_{w,d}(k)$.
Likewise,
Eq.~\eqref{atmftheta} normalizes the incoming messages attached the author index $a$ in terms of the topic index $k$ to
make outgoing messages comparable for different authors $a \in \mathbf{a}_d$.
Similar to~\eqref{v2f3},
we derive the message passing $\mu_{f_{\theta_a} \rightarrow z^a_{w,d}}$ through adding all incoming messages evaluated by the factor function~\eqref{atmftheta}.

Multiplying two messages from factors $\theta_{a \in \mathbf{a}_d}$ and $\phi_w$ yields the message update equation as follows,
\begin{align} \label{atmmessage}
\mu^a_{w,d}(k) \propto \frac{\boldsymbol{\mu}_{-w,-d}^a(k) + \alpha}
{\sum_k [\boldsymbol{\mu}_{-w,-d}^a(k) + \alpha]}
\times \frac{\boldsymbol{\mu}_{w,-d}(k) + \beta}{\sum_w [\boldsymbol{\mu}_{w,-d}(k) + \beta]}.
\end{align}
Notice that the $|\mathbf{a}_d|\times K$-tuple message
$\mu^a_{w,d}(k), a \in \mathbf{a}_d$ is normalized in terms of all
combinations of $\{a,k\}, a \in \mathbf{a}_d, 1 \le k \le K$. Based
on the normalized messages, the author-specific topic proportion
$\theta_a(k)$ can be estimated from the sum of all incoming messages
including $\mu^a_{w,d}$ evaluated by the factor function
$f_{\theta_a}$ as follows,
\begin{align} \label{atmtheta}
\theta_a(k) = \frac{\boldsymbol{\mu}^a_{\cdot,\cdot}(k) + \alpha}{\sum_k [\boldsymbol{\mu}^a_{\cdot,\cdot}(k) + \alpha]}.
\end{align}

As a summary,
Fig.~\ref{BPATM} shows the synchronous BP algorithm for learning ATM.
The difference between Fig.~\ref{BP} and Fig.~\ref{BPATM} is that Fig.~\ref{BP} considers the author index $a$ as the label for each word.
At each iteration,
the computational complexity is $\mathcal{O}(KDW_dA_dT)$,
where $A_d$ is the average number of authors per document.

\section{Belief Propagation for RTM} \label{s4}

\begin{figure*}[t]
\centering
\includegraphics[width=0.7\linewidth]{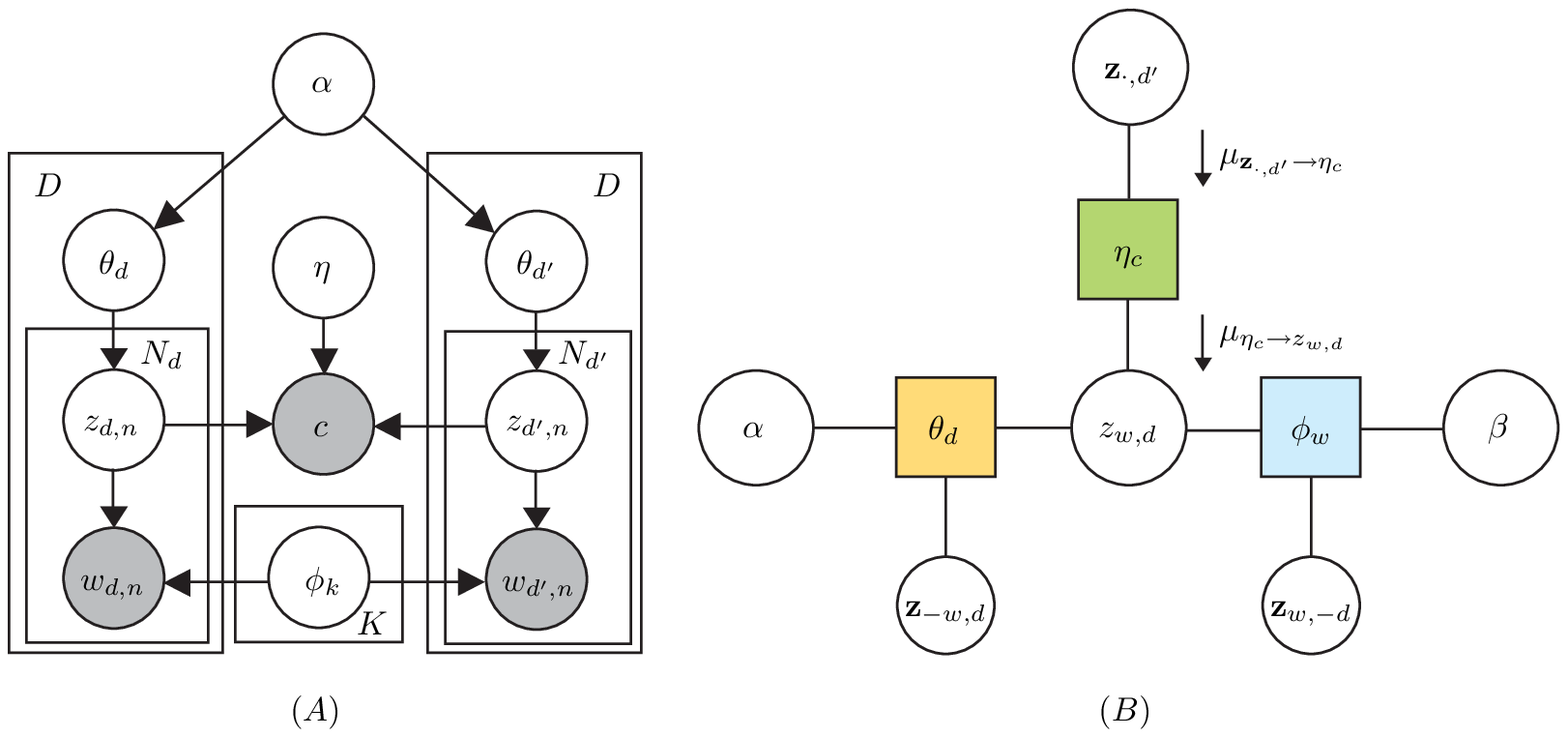}
\caption{(A) The three-layer graphical representation~\cite{Chang:10} and (B) two-layer factor graph of RTM.
}
\label{rtm}
\end{figure*}

Network data,
such as citation and coauthor networks of documents~\cite{Zeng:09,Zeng:10},
tag networks of documents and images~\cite{Zeng:11},
hyperlinked networks of web pages,
and social networks of friends,
exist pervasively in data mining and machine learning.
The probabilistic relational topic modeling of network data can provide both useful predictive models and descriptive statistics~\cite{Chang:10}.

In Fig.~\ref{rtm}A,
relational topic models (RTM)~\cite{Chang:10} represent entire document topics by the mean value of the document topic proportions,
and use Hadamard product of mean values $\overline{z}_d \circ \overline{z}_{d'}$ from two linked documents $\{d,d'\}$ as link features,
which are learned by the generalized linear model (GLM) $\eta$ to generate the observed binary citation link variable $c=1$.
Besides,
all other parts in RTM remain the same as LDA.

We transform Fig.~\ref{rtm}A to the factor graph Fig.~\ref{rtm}B by absorbing the observed link index $c \in \mathbf{c}, 1 \le c \le C$ as the index of the factor $\eta_c$.
Each link index connects a document pair $\{d,d'\}$,
and the factor $\eta_c$ connects word topic labels $z_{w,d}$ and $\mathbf{z}_{\cdot,d'}$ of the document pair.
Besides encoding the topic smoothness,
RTM explicitly describes the topic structural dependencies between the pair of linked documents $\{d,d'\}$ using the factor function $f_{\eta_c}(\cdot)$.

\subsection{Inference and Parameter Estimation}

\begin{figure}[t]
\centering
\includegraphics[width=1\linewidth]{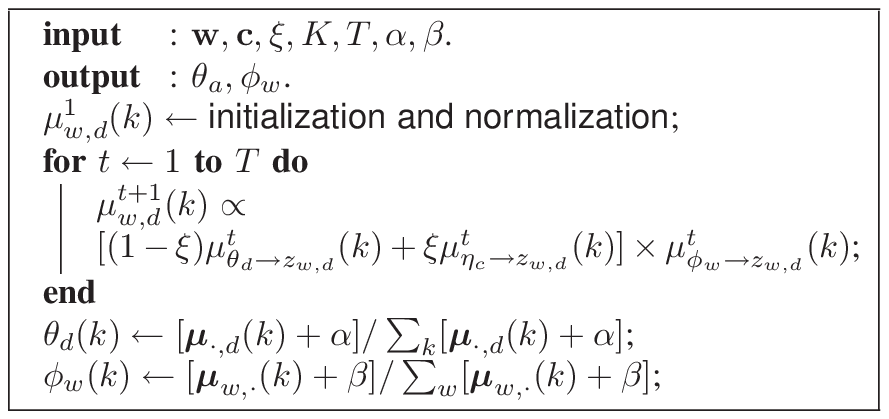}
\caption{The synchronous BP for RTM.}
\label{BPRTM}
\end{figure}

In Fig.~\ref{rtm},
the messages from the factors $\theta_d$ and $\phi_w$ to the variable $z_{w,d}$ are the same as LDA in~\eqref{v2f3} and~\eqref{v2f4}.
Thus,
we only need to derive the message passing equation from the factor $\eta_c$ to the variable $z_{w,d}$.

We design the factor function $f_{\eta_c}(\cdot)$ for linked documents as follows,
\begin{align} \label{eta}
f_{\eta_c}(k|k') = \frac{\sum_{\{d,d'\}}\boldsymbol{\mu}_{\cdot,d}(k)\boldsymbol{\mu}_{\cdot,d'}(k')}{\sum_{\{d,d'\},k'}\boldsymbol{\mu}_{\cdot,d}(k)\boldsymbol{\mu}_{\cdot,d'}(k')},
\end{align}
which depicts the likelihood of topic label $k$ assigned to the
document $d$ when its linked document $d'$ is associated with the
topic label $k'$. Notice that the designed factor function does not
follow the GLM for link modeling in the original RTM~\cite{Chang:10}
because the GLM makes inference slightly more complicated. However,
similar to the GLM, Eq.~\eqref{eta} is also able to capture the
topic interactions between two linked documents $\{d,d'\}$ in
document networks. Instead of smoothness prior encoded by factor
functions~\eqref{ftheta} and~\eqref{fphi}, it describes arbitrary
topic dependencies $\{k,k'\}$ of linked documents $\{d,d'\}$.

Based on the factor function~\eqref{eta},
we resort to the sum-product algorithm to calculate the message,
\begin{align}
\mu_{\eta_c \rightarrow z_{w,d}}(k) = \sum_{d'}\sum_{k'}f_{\eta_c}(k|k')\boldsymbol{\mu}_{\cdot,d'}(k'),
\end{align}
where we use the sum rather than the product of messages from all linked documents $d'$ to avoid arithmetic underflow.
The standard sum-product algorithm requires the product of all messages from factors to variables.
However,
in practice,
the direct product operation cannot balance the messages from different sources.
For example,
the message $\mu_{{\theta_d} \rightarrow z_{w,d}}$ is from the neighboring words within the same document $d$,
while the message $\mu_{\eta_c \rightarrow z_{w,d}}$ is from all linked documents $d'$.
If we pass the product of these two types of messages,
we cannot distinguish which one influences more on the topic label $z_{w,d}$.
Hence,
we use the weighted sum of two types of messages,
\begin{align} \label{rtmmessage}
\mu(z_{w,d}=k) \propto& [(1-\xi)\mu_{{\theta_d} \rightarrow z_{w,d}}(k) \notag \\
&+ \xi\mu_{\eta_c \rightarrow z_{w,d}}(k)] \times
\mu_{\phi_w \rightarrow z_{w,d}}(k),
\end{align}
where $\xi \in [0,1]$ is the weight to balance two messages $\mu_{{\theta_d} \rightarrow z_{w,d}}$ and $\mu_{\eta_c \rightarrow z_{w,d}}$.
When there are no link information $\xi = 0$,
Eq.~\eqref{rtmmessage} reduces to~\eqref{message} so that RTM reduces to LDA.
Fig.~\ref{BPRTM}
shows the synchronous BP algorithm for learning RTM.
Given the inferred messages,
the parameter estimation equations remain the same as~\eqref{thetad} and~\eqref{phiw}.
The computational complexity at each iteration is $\mathcal{O}(K^2CDW_dT)$,
where $C$ is the total number of links in the document network.

\section{Experiments} \label{s5}

\begin{table}[t]
\centering
\caption{Summarization of four document data sets}
\begin{tabular}{|c|c|c|c|c|c|c|} \hline
Data sets   &$D$         &$A$      &$W$     &$C$      &$N_d$    &$W_d$     \\ \hline \hline
CORA        &$2410$      &$2480$   &$2961$  &$8651$   &$57$     &$43$      \\
MEDL        &$2317$      &$8906$   &$8918$  &$1168$   &$104$    &$66$      \\
NIPS        &$1740$      &$2037$   &$13649$ &$-$      &$1323$   &$536$     \\
BLOG        &$5177$      &$-$      &$33574$ &$1549$   &$217$    &$149$     \\ \hline
\end{tabular}
\label{datasets}
\end{table}

We use four large-scale document data sets:
\begin{enumerate}
\item
CORA~\cite{McCallum:00} contains abstracts from the CORA research paper search engine in machine learning area,
where the documents can be classified into $7$ major categories.
\item
MEDL~\cite{Zhu:09a} contains abstracts from the MEDLINE biomedical paper search engine,
where the documents fall broadly into $4$ categories.
\item
NIPS~\cite{Globerson:07} includes papers from the conference ``Neural Information Processing Systems",
where all papers are grouped into $13$ categories.
NIPS has no citation link information.
\item
BLOG~\cite{Eisenstein:10} contains a collection of political blogs on the subject of American politics in the year 2008.
where all blogs can be broadly classified into $6$ categories.
BLOG has no author information.
\end{enumerate}
Table~\ref{datasets} summarizes the statistics of four data sets,
where $D$ is the total number of documents,
$A$ is the total number of authors,
$W$ is the vocabulary size,
$C$ is the total number of links between documents,
$N_d$ is the average number of words per document,
and $W_d$ is the average vocabulary size per document.

\subsection{BP for LDA}

\begin{figure*}[t]
\centering
\includegraphics[width=0.8\linewidth]{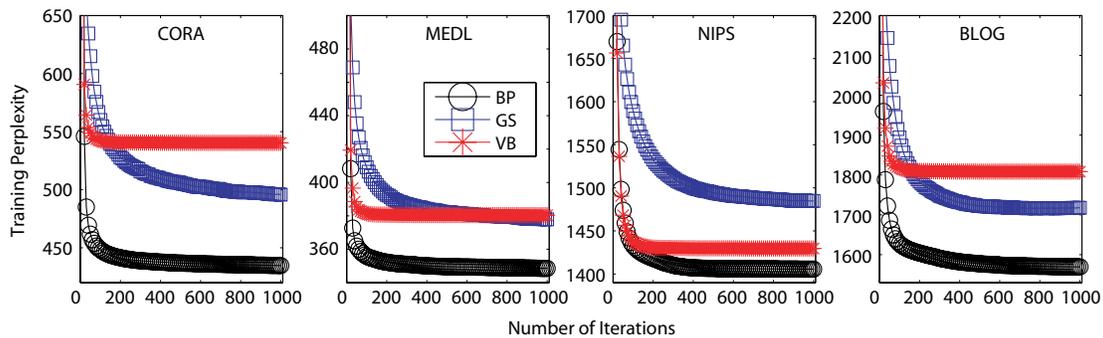}
\caption{Training perplexity as a function of number of iterations when $K=50$ on CORA, MEDL, NIPS and BLOG.
}
\label{convergence}
\end{figure*}

We compare BP with two commonly-used LDA learning algorithms such as VB~\cite{Blei:03}
(Here we use Blei's implementation of digamma functions)\footnote{\url{http://www.cs.princeton.edu/~blei/lda-c/index.html}}
and GS~\cite{Griffiths:04}\footnote{\url{http://psiexp.ss.uci.edu/research/programs_data/toolbox.htm}} under the same fixed hyperparameters $\alpha = \beta = 0.01$.
We use MATLAB C/C++ MEX-implementations for all these algorithms,
and carry out experiments on a common PC with CPU $2.4$GHz and RAM $4$G.
With the goal of repeatability,
we have made our source codes and data sets publicly available~\cite{Zeng:12}.

To examine the convergence property of BP,
we use the entire data set as the training set,
and calculate the training perplexity~\cite{Blei:03} at every $10$ iterations in the total of $1000$ training iterations from the same initialization.
Fig.~\ref{convergence} shows that the training perplexity of BP generally decreases rapidly as the number of training iterations increases.
In our experiments,
BP on average converges with the number of training iterations $T \approx 170$
when the difference of training perplexity between two successive iterations is less than one.
Although this paper does not theoretically prove that BP will definitely converge to the fixed point,
the resemblance among VB, GS and BP in the subsection~\ref{discussion} implies
that there should be the similar underlying principle that ensures BP to converge on general sparse word vector space in real-world applications.
Further analysis reveals that BP on average uses more number of training iterations until convergence than VB ($T \approx 100$)
but much less number of training iterations than GS ($T \approx 300$) on the four data sets.
The fast convergence rate is a desirable property as far as the online~\cite{Hoffman:10} and
distributed~\cite{Zhai:11} topic modeling for large-scale corpus are concerned.

\begin{figure*}[t]
\centering
\includegraphics[width=0.8\linewidth]{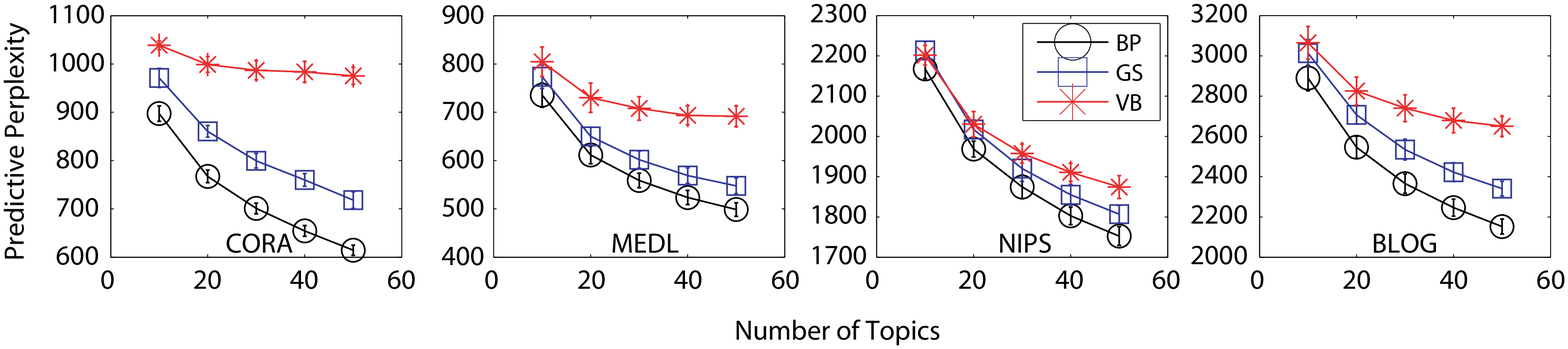}
\caption{Predictive perplexity as a function of number of topics on CORA, MEDL, NIPS and BLOG.
}
\label{perplda}
\end{figure*}

The predictive perplexity for the unseen test set is computed as follows~\cite{Blei:03,Asuncion:09}.
To ensure all algorithms to achieve the local optimum,
we use the $1000$ training iterations to estimate $\phi$ on the training set from the same initialization.
In practice,
this number of training iterations is large enough for convergence of all algorithms in Fig.~\ref{convergence}.
We randomly partition each document in the test set into $90\%$ and $10\%$ subsets.
We use $1000$ iterations of learning algorithms to estimate $\theta$ from the same initialization while holding $\phi$ fixed on the $90\%$ subset,
and then calculate the predictive perplexity on the left $10\%$ subset,
\begin{align}
\label{perplexity}
\mathcal{P} = \exp\Bigg\{-\frac{\sum_{w,d}
x_{w,d}^{10\%}\log\big[\sum_{k}\theta_d(k)\phi_w(k)\big]}
{\sum_{w,d} x_{w,d}^{10\%}}\Bigg\},
\end{align}
where $x_{w,d}^{10\%}$ denotes word counts in the $10\%$ subset.
Notice that the perplexity~\eqref{perplexity} is based on the
marginal probability of the word topic label $\mu_{w,d}(k)$
in~\eqref{approx}.

\begin{figure*}[t]
\centering
\includegraphics[width=0.8\linewidth]{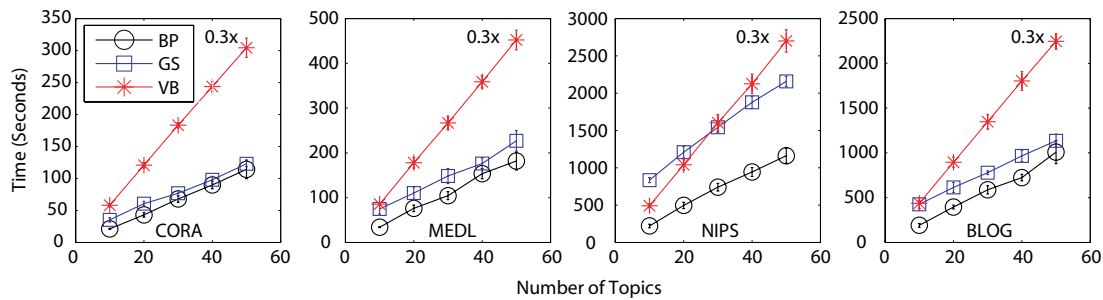}
\caption{Training time as a function of number of topics on CORA, MEDL, NIPS and BLOG.
For VB,
it shows $0.3$ times of the real learning time denoted by $0.3$x.
}
\label{timelda}
\end{figure*}

Fig.~\ref{perplda} shows the predictive perplexity (average $\pm$ standard deviation) from five-fold cross-validation for different topics,
where the lower perplexity indicates the better generalization ability for the unseen test set.
Consistently,
BP has the lowest perplexity for different topics on four data sets,
which confirms its effectiveness for learning LDA.
On average,
BP lowers around $11\%$ than VB and $6\%$ than GS in perplexity.
Fig.~\ref{timelda} shows that BP uses less training time than both VB and GS.
We show only $0.3$ times of the real training time of VB because of time-consuming digamma functions.
In fact,
VB runs as fast as BP if we remove digamma functions.
So,
we believe that it is the digamma functions that slow down VB in learning LDA.
BP is faster than GS because it computes messages for word indices.
The speed difference is largest on the NIPS set due to its largest ratio $N_d/W_d=2.47$ in Table~\ref{datasets}.
Although VB converges rapidly attributed to digamma functions,
it often consumes triple more training time.
Therefore,
BP on average enjoys the highest efficiency for learning LDA with regard to the balance of convergence rate and training time.

\begin{figure*}[t]
\centering
\includegraphics[width=0.8\linewidth]{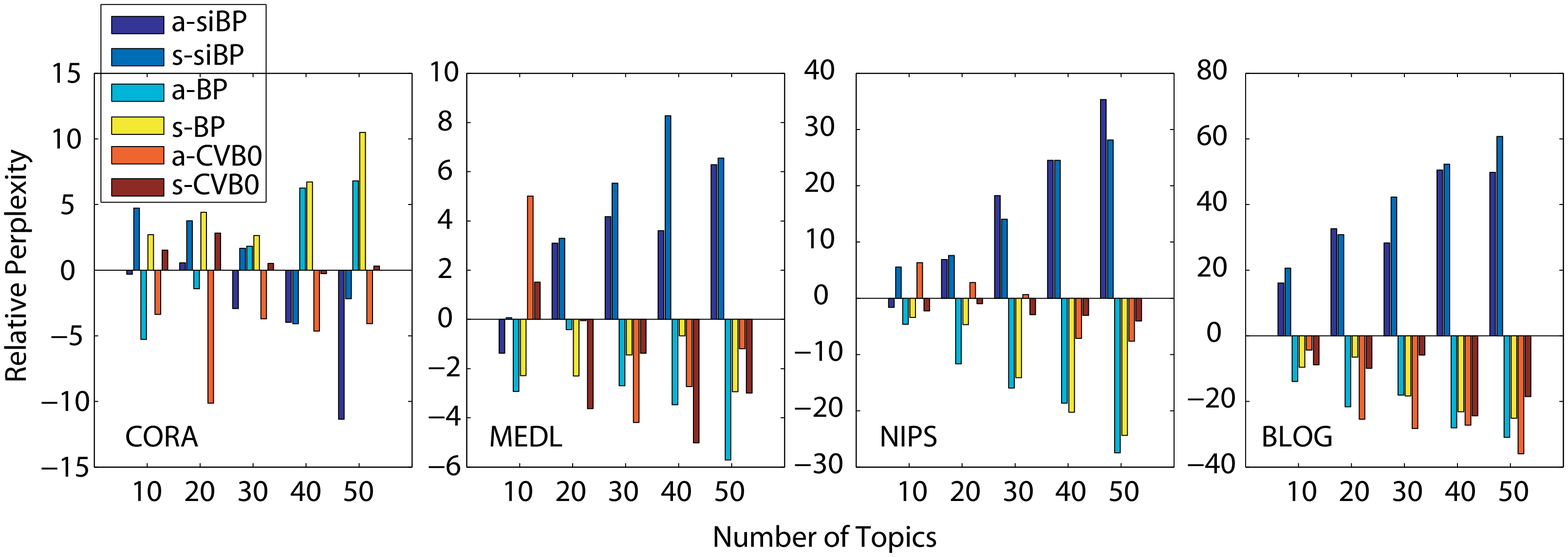}
\caption{Relative predictive perplexity as a function of number of topics on CORA, MEDL, NIPS and BLOG.
}
\label{bpall}
\end{figure*}

We also compare six BP implementations such as
siBP, BP and CVB0~\cite{Asuncion:09} using both synchronous and asynchronous update schedules.
We name three synchronous implementations as s-BP, s-siBP and s-CVB0,
and three asynchronous implementations as a-BP, a-siBP and a-CVB0.
Because these six belief propagation implementations produce comparable perplexity,
we show the relative perplexity that subtracts the mean value of six implementations in Fig.~\ref{bpall}.
Overall,
the asynchronous schedule gives slightly lower perplexity than synchronous schedule because it passes messages faster and more efficiently.
Except on CORA set,
siBP generally provides the highest perplexity because it introduces subtle biases in computing messages at each iteration.
The biased message will be propagated and accumulated leading to inaccurate parameter estimation.
Although the proposed BP achieves lower perplexity than CVB0 on NIPS set,
both of them work comparably well on other sets.
But BP is much faster because it computes messages over word indices.
The comparable results also confirm our assumption that topic modeling
can be efficiently performed on word indices instead of tokens.

\begin{figure*}[t]
\centering
\includegraphics[width=0.8\linewidth]{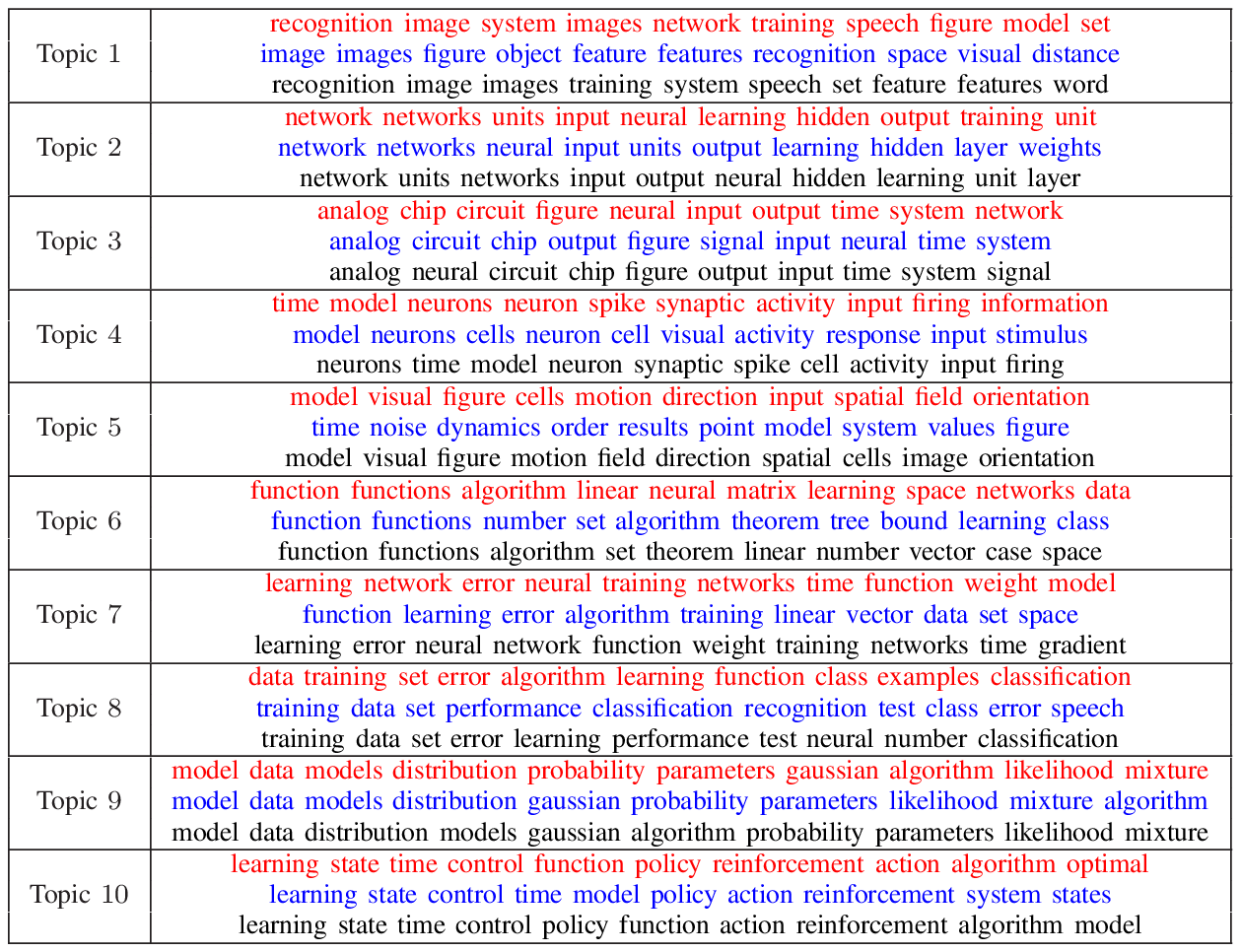}
\caption{Top ten words of $K=10$ topics of VB (first line), GS (second line), and BP (third line) on NIPS.
}
\label{topic}
\end{figure*}

To measure the interpretability of a topic model,
the {\em word intrusion} and {\em topic intrusion} are proposed to involve subjective judgements~\cite{Chang:09b}.
The basic idea is to ask volunteer subjects to identify the number of word intruders in the topic as well as the topic intruders in the document,
where intruders are defined as inconsistent words or topics based on prior knowledge of subjects.
Fig.~\ref{topic} shows the top ten words of $K=10$ topics inferred by VB, GS and BP algorithms on NIPS set.
We find no obvious difference with respect to word intrusions in each topic.
Most topics share the similar top ten words but with different ranking orders.
Despite significant perplexity difference,
the topics extracted by three algorithms remains almost the same interpretability at least for the top ten words.
This result coincides with~\cite{Chang:09b} that the lower perplexity may not enhance interpretability of inferred topics.

\begin{figure}[t]
\centering
\includegraphics[width=0.8\linewidth]{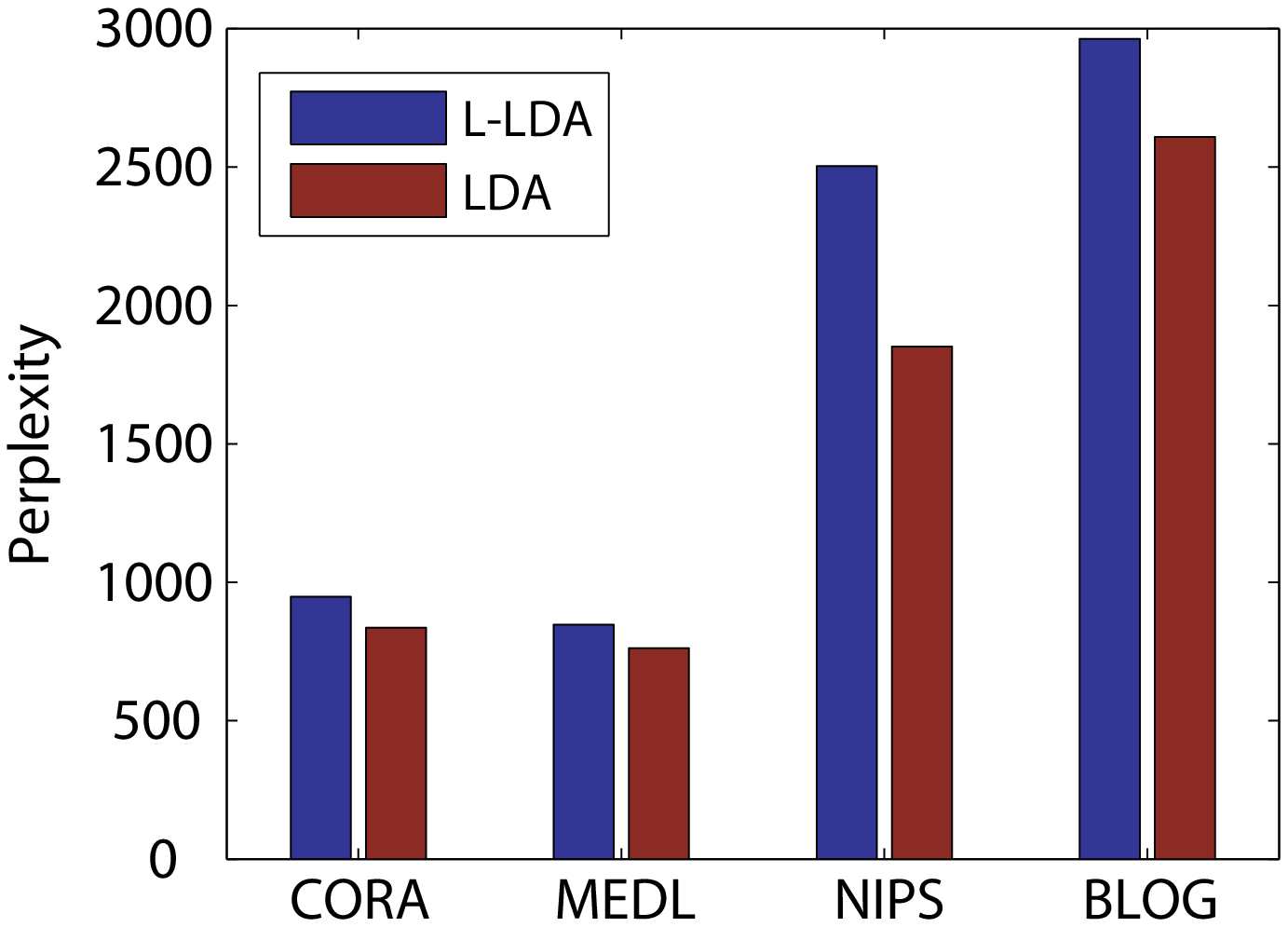}
\caption{Perplexity of L-LDA and LDA on four data sets.
}
\label{lalda}
\end{figure}

Similar phenomenon has also been observed in MRF-based image labeling problems~\cite{Tappen:03}.
Different MRF inference algorithms such as graph cuts and BP often yield comparable results.
Although one inference method may find more optimal MRF solutions,
it does not necessarily translate into better performance compared to the ground-truth.
The underlying hypothesis is that the ground-truth labeling configuration is often less optimal than solutions produced by inference algorithms.
For example,
if we manually label the topics for a corpus,
the final perplexity is often higher than that of solutions returned by VB, GS and BP.
For each document,
LDA provides the equal number of topics $K$ but the ground-truth often uses the unequal number of topics to explain the observed words,
which may be another reason why the overall perplexity of learned LDA is often lower than that of the ground-truth.
To test this hypothesis,
we compare the perplexity of labeled LDA (L-LDA)~\cite{Ramage:09} with LDA in Fig.~\ref{lalda}.
L-LDA is a supervised LDA that restricts the hidden topics as the observed class labels of each document.
When a document has multiple class labels,
L-LDA automatically assigns one of the class labels to each word index.
In this way,
L-LDA resembles the process of manual topic labeling by human,
and its solution can be viewed as close to the ground-truth.
For a fair comparison,
we set the number of topics $K=7,4,13,6$ of LDA for CORA, MEDL, NIPS and BLOG according to the number of document categories in each set.
Both L-LDA and LDA are trained by BP using $500$ iterations from the same initialization.
Fig.~\ref{lalda} confirms that L-LDA produces higher perplexity than LDA,
which partly supports that the ground-truth often yields the higher perplexity than the optimal solutions of LDA inferred by BP.
The underlying reason may be that the three topic modeling rules encoded by LDA are still too simple to capture
human behaviors in finding topics.

Under this situation,
improving the formulation of topic models such as LDA is better than improving inference algorithms
to enhance the topic modeling performance significantly.
Although the proper settings of hyperparameters can make the predictive perplexity comparable
for all state-of-the-art approximate inference algorithms~\cite{Asuncion:09},
we still advocate BP because it is faster and more accurate than both VB and GS,
even if they all can provide comparable perplexity and interpretability under the proper settings of hyperparameters.

\subsection{BP for ATM}

\begin{figure*}[t]
\centering
\includegraphics[width=0.8\linewidth]{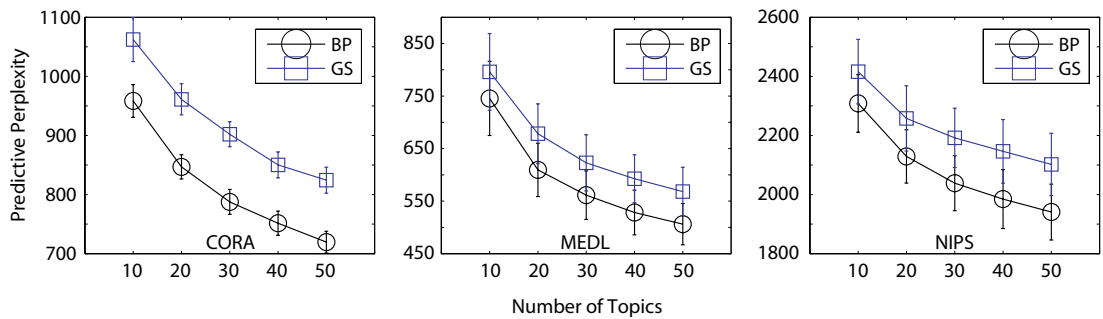}
\caption{Predictive perplexity as a function of number of topics for ATM on CORA, MEDL and NIPS.
}
\label{perpatm}
\end{figure*}

The GS algorithm for learning ATM is implemented in the MATLAB topic modeling toolbox.\footnote{\url{http://psiexp.ss.uci.edu/research/programs_data/toolbox.htm}}
We compare BP and GS for learning ATM based on $500$ iterations on training data.
Fig.~\ref{perpatm} shows the predictive perplexity (average $\pm$ standard deviation) from five-fold cross-validation.
On average,
BP lowers $12\%$ perplexity than GS,
which is consistent with Fig.~\ref{perplda}.
Another possible reason for such improvements may be our assumption that
all coauthors of the document account for the word topic label using multinomial instead of uniform probabilities.

\subsection{BP for RTM}

\begin{figure*}[t]
\centering
\includegraphics[width=0.8\linewidth]{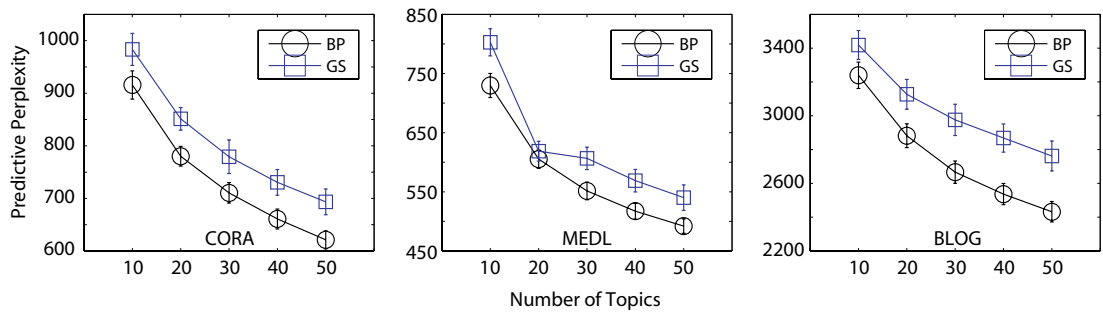}
\caption{Predictive perplexity as a function of number of topics for RTM on CORA, MEDL and BLOG.
}
\label{perprtm}
\end{figure*}

\begin{figure*}[t]
\centering
\includegraphics[width=0.8\linewidth]{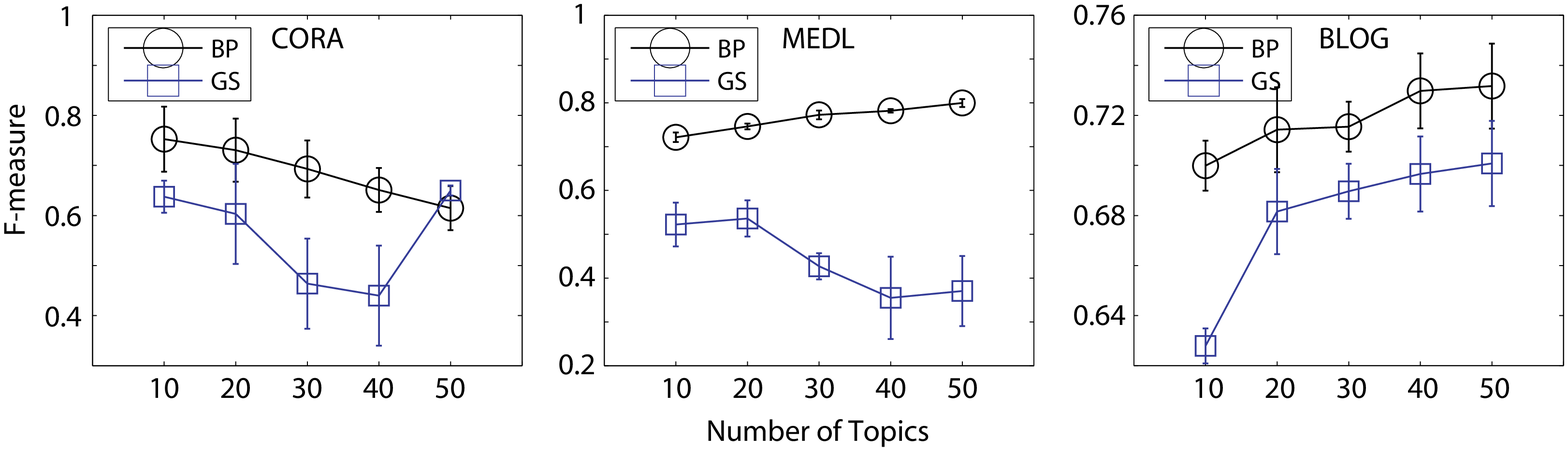}
\caption{F-measure of link prediction as a function of number of topics on CORA, MEDL and BLOG.
}
\label{link}
\end{figure*}

The GS algorithm for learning RTM is implemented in the R package.\footnote{\url{http://cran.r-project.org/web/packages/lda/}}
We compare BP with GS for learning RTM using the same $500$ iterations on training data set.
Based on the training perplexity,
we manually set the weight $\xi=0.15$ in Fig.~\ref{BPRTM} to achieve the overall superior performance on four data sets.

Fig.~\ref{perprtm} shows predictive perplexity (average $\pm$ standard deviation) on five-fold cross-validation.
On average,
BP lowers $6\%$ perplexity than GS.
Because the original RTM learned by GS is inflexible to balance information from different sources,
it has slightly higher perplexity than LDA (Fig.~\ref{perplda}).
To circumvent this problem,
we introduce the weight $\xi$ in~\eqref{rtmmessage} to balance two types of messages,
so that the learned RTM gains lower perplexity than LDA.
Future work will estimate the balancing weight $\xi$
based on the feature selection or MRF structure learning techniques.

We also examine the link prediction performance of RTM.
We define the link prediction as a binary classification problem.
As with~\cite{Chang:10},
we use the Hadmard product of a pair of document topic proportions as the link feature,
and train an SVM~\cite{Chang:11} to decide if there is a link between them.
Notice that the original RTM~\cite{Chang:10} learned by the GS algorithm uses the GLM to predict links.
Fig.~\ref{link} compares the F-measure (average $\pm$ standard deviation) of link prediction on five-fold cross-validation.
Encouragingly,
BP provides significantly $15\%$ higher F-measure over GS on average.
These results confirm the effectiveness of BP for capturing accurate topic structural dependencies in document networks.

\section{Conclusions} \label{s6}

First,
this paper has presented the novel factor graph representation of LDA within the MRF framework.
Not only does MRF solve topic modeling as a labeling problem,
but also facilitate BP algorithms for approximate inference and parameter estimation in three steps:
\begin{enumerate}
\item
First,
we absorb $\{w,d\}$ as indices of factors,
which connect hidden variables such as topic labels in the neighborhood system.
\item
Second,
we design the proper factor functions to encourage or penalize different local topic labeling configurations in the neighborhood system.
\item
Third,
we develop the approximate inference and parameter estimation algorithms within the message passing framework.
\end{enumerate}
The BP algorithm is easy to implement,
computationally efficient,
faster and more accurate than other two approximate inference methods
like VB~\cite{Blei:03} and GS~\cite{Griffiths:04} in several topic modeling tasks of broad interests.
Furthermore,
the superior performance of BP algorithm for learning ATM~\cite{Rosen-Zvi:04} and RTM~\cite{Chang:10}
confirms its potential effectiveness in learning other LDA extensions.

Second,
as the main contribution of this paper,
the proper definition of neighborhood systems as well as the design of factor functions can interpret the three-layer LDA by the two-layer MRF
in the sense that they encode the same joint probability.
Since the probabilistic topic modeling is essentially a word annotation paradigm,
the opened MRF perspective may inspire us to use other MRF-based image segmentation~\cite{Shi:00}
or data clustering algorithms~\cite{Frey:07} for LDA-based topic models.

Finally,
the scalability of BP is an important issue in our future work.
As with VB and GS,
the BP algorithm has a linear complexity with the number documents $D$ and the number of topics $K$.
We may extend the proposed BP algorithm for online~\cite{Hoffman:10} and distributed~\cite{Zhai:11} learning of LDA,
where the former incrementally learns parts of $D$ documents in data streams and the latter learns parts of $D$ documents on distributed computing units.
Since the $K$-tuple message is often sparse~\cite{Porteous:09},
we may also pass only salient parts of the $K$-tuple messages or only update those informative parts of messages
at each learning iteration to speed up the whole message passing process.

\section*{Acknowledgements} \label{s7}

This work is supported by NSFC (Grant No. 61003154),
the Shanghai Key Laboratory of Intelligent Information Processing, China (Grant No. IIPL-2010-009),
and a grant from Baidu.

\bibliographystyle{IEEEtran}
\bibliography{IEEEabrv,TMBP}

\end{document}